\pdfoutput=1
\documentclass[letterpaper, 10 pt, conference]{ieeeconf}
\usepackage{graphicx}
\usepackage{calrsfs}
\usepackage{algorithm}
\usepackage{algorithmic}
\usepackage{amssymb}
\usepackage[colorlinks]{hyperref}
\usepackage{xcolor}
\DeclareMathAlphabet{\pazocal}{OMS}{zplm}{m}{n}
\usepackage[font={footnotesize}]{caption}

\newcommand{\lfpwebsite}{https://learning-from-play.github.io}
\newcommand{\lfpvideos}{\lfpwebsite/assets/mp4}
\newcommand{\website}{https://learning-to-play.github.io}
\newcommand{\ltpvideos}{\website/videos}

\IEEEoverridecommandlockouts  % This command is only needed if you want to use the \thanks command

\title{\LARGE \bf
Learning to Play by Imitating Humans
}

\author{Rostam Dinyari$^{1}$, Pierre Sermanet$^{2}$ and Corey Lynch$^{2}$% <-this % stops a space
\\*\{\tt\small {rostam, sermanet, coreylynch\}@google.com}
\thanks{$^{1}$Google Cloud}%
\thanks{$^{2}$Robotics at Google (\href{http://g.co/robotics}{g.co/robotics})}%
}

\begin{document}

\maketitle
\thispagestyle{empty}
\pagestyle{empty}

%%%%%%%%%%%%%%%%%%%%%%%%%%%%%%%%%%%%%%%%%%%%%%%%%%%%%%%%%%%%%%%%%%%%%%%%%%%%%%%%
\begin{abstract}

Acquiring multiple skills has commonly involved collecting a large number of expert demonstrations per task or engineering custom reward functions. Recently it has been shown that it is possible to acquire a diverse set of skills by self-supervising control on top of human teleoperated play data. Play is rich in state space coverage and a policy trained on this data can generalize to specific tasks at test time outperforming policies trained on individual expert task demonstrations. In this work, we explore the question of whether robots can learn to play to autonomously generate play data that can ultimately enhance performance. By training a behavioral cloning policy on a relatively small quantity of human play, we autonomously generate a large quantity of cloned play data that can be used as additional training. We demonstrate that a general purpose goal-conditioned policy trained on this augmented dataset substantially outperforms one trained only with the original human data on 18 difficult user-specified manipulation tasks in a simulated robotic tabletop environment. A video example of a robot imitating human play can be seen here: \href{\ltpvideos/undirected_play1.mp4}{\ltpvideos/undirected\_play1.mp4}    

\end{abstract}

%%%%%%%%%%%%%%%%%%%%%%%%%%%%%%%%%%%%%%%%%%%%%%%%%%%%%%%%%%%%%%%%%%%%%%%%%%%%%%%%
\section{INTRODUCTION}

Building generalist robots remains a difficult long term problem in robotics and the attempts to train them have involved considerable human effort. Training a robot that can perform many different tasks generally involves using large datasets to train high capacity machine learning models \cite{kalashnikov, akkaya, nicolescu, pastor, kober, deisenroth, kroemer}. Often these datasets consist of labeled and segmented expert demonstrations of the tasks to be performed by the robot. These demonstrations are then used to train a model that imitates these behaviors. This process results in specialist robots that can perform specific tasks such as grasping \cite{hsiao, steil, schaal, pathak}, dexterous manipulations \cite{antotsiou}, or locomotion \cite{ratliff}. Collecting rich datasets that can train a robot subtle details of the desired tasks is labor intensive \cite{pastor}. Unfortunately autonomous methods of collecting training examples rarely result in datasets that are rich and often require using other techniques such as engineering reward functions \cite{rajeswaran, akkaya}.

\begin{figure}[t]
\centering
\includegraphics[width=0.48\textwidth]{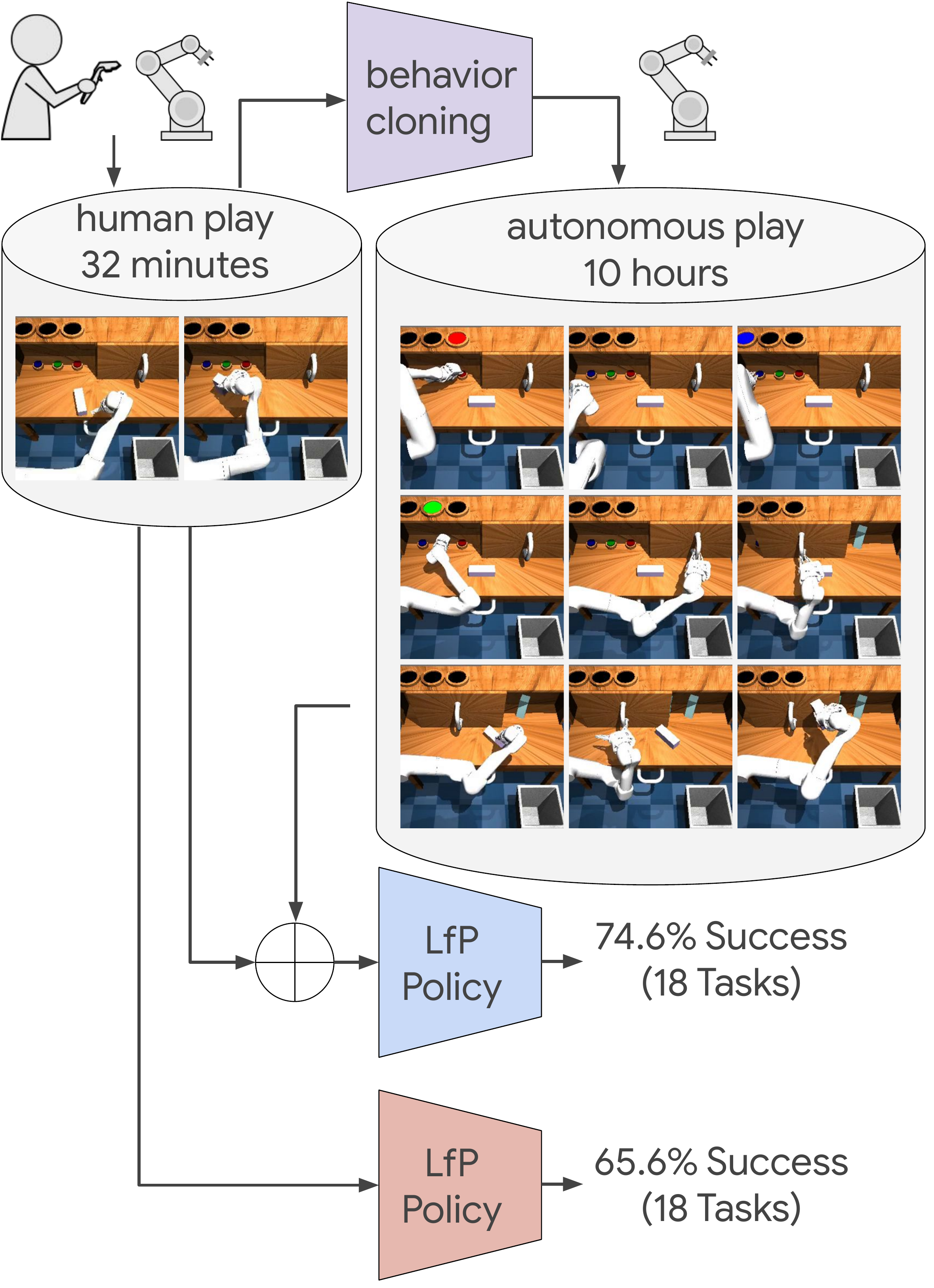}
\caption{Learning to generate useful play data by imitating humans. We start by collecting a small quantity of human teleoperated play data. We use the human play data to train a BC policy. This policy imitates play behavior to generate a new set of autonomous (cloned) play data. The collected autonomous play data is added to the human play data to train an LfP policy and evaluated on 18 complex user-specified tasks.}
\label{fig:play_cloning}
\end{figure}

Recently self-supervision on top of teleoperated human play data has been proposed as a way to learn goal-directed control without the need to manually create a labeled dataset for each task \cite{lynch}. In this setting, a human teleoperates the robot for an extended period of time, engaging in as many manipulation behaviors as they can think of, deliberately covering an environments available state space — see videos of the real \href{\lfpvideos/play_data516x360.mp4}{play data} collected for \cite{lynch}. For example, a human operator may teleoperate a robot to open a drawer, move a sliding door, rotate a block clockwise, and press a button — whatever satisfies their curiosity. With a simple relabeling scheme, Lynch et al. \cite{lynch} convert a few hours of play into many diverse, short-horizon, hindsight goal-relabeled task examples for use in a goal conditioned imitation learning context. This enables the training of a single general purpose goal-conditioned policy, capable of generalizing in zero shot to 18 user-specified visual manipulation tasks, exceeding the performance of conventional behavioral cloning (BC) policies trained on individual task expert demonstration datasets.

\begin{figure}[t]
\centering
\includegraphics[width=0.48\textwidth]{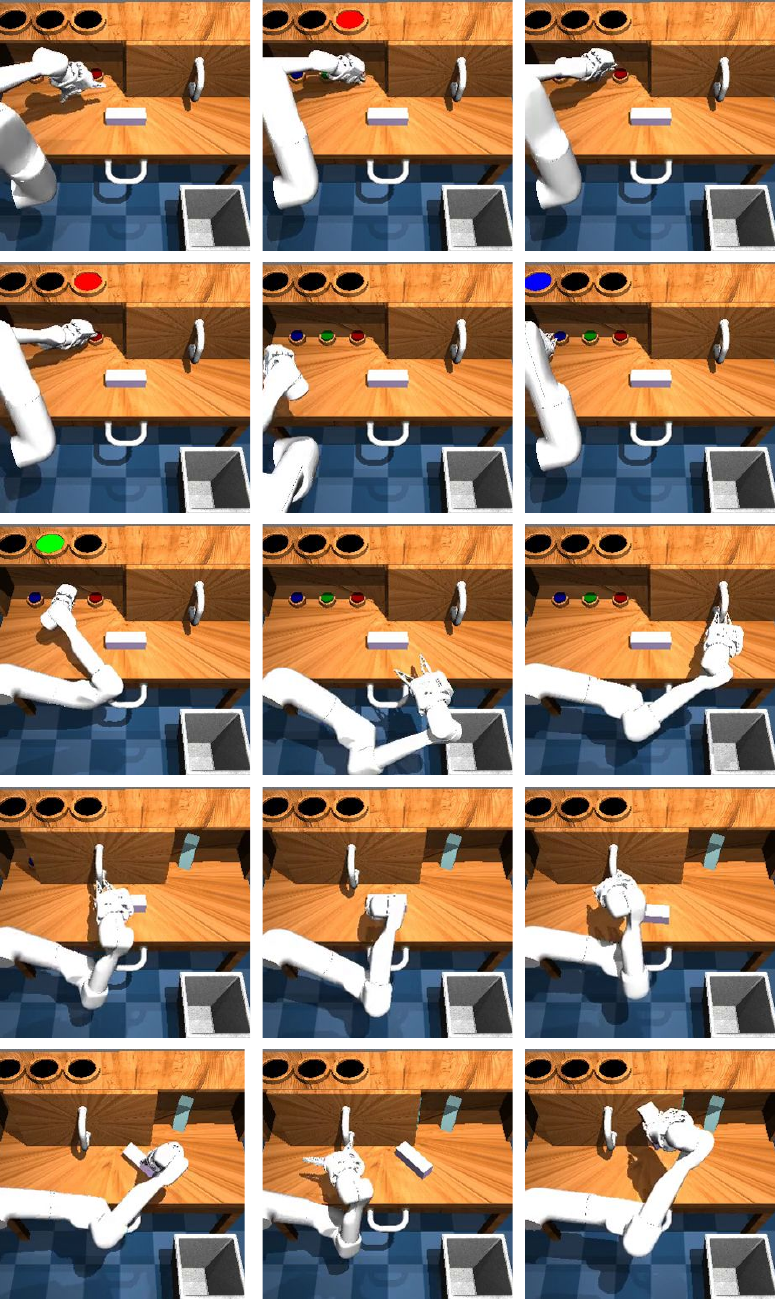}
\caption{An example of autonomous robot play, generated using a policy trained to imitate how people play. A robot presses a red button, releases the red button, presses the red button again, presses a blue button, presses a green button, moves a sliding door to the left, sweeps a rectangular block, moves the sliding door, and picks up the block. This cheap, unstructured, and  interaction-rich data can be used to help train precise goal-directed manipulation.}
\label{fig:cloned_play_sequence}
\end{figure}

Although human teleoperated play can be collected in large quantities cheaply, it still requires a human to be providing the play. A more scalable setup is one in which robots collect this kind of play for us. In this paper, we explore whether we can train a robot to play like a person — the cheapest way of expanding a high-coverage play dataset. To explore this question, we introduce a simple and scalable method — see Fig. \ref{fig:play_cloning} — for learning to play: we treat small quantities of cheaply collected human teleoperated play — here also referred to as human play for short — as optimal behavior describing of \emph{how to play}, then train an imitation learning policy on top via supervised learning. Given a trained policy, large amounts of additional manipulation-rich interaction data can be generated autonomously, by taking samples from the learned policy in the complex environment. As we see in Fig. \ref{fig:cloned_play_sequence}, this results in useful functional behaviors, e.g. opening doors, opening drawers, picking up objects on a table, pressing buttons, etc. Crucially, via the relabeling in \cite{lynch}, this cloned autonomous play data can be used in addition to human play data to learn general goal-conditioned policies. We demonstrate that when supplementing the human play data with the cloned play data produced by our method, goal-conditioned policies trained on top of the combined dataset exhibit substantially better downstream performance on 18 complex user-specified manipulation tasks than when training on human play alone.

\section{RELATED WORK}

A dominant way to create learning robots is to use some form of supervised learning, for example, learning from demonstrations \cite{nicolescu, pastor} or reinforcement learning \cite{kober, deisenroth, kalashnikov, akkaya}. These methods require considerable human effort, for example, to acquire specialized training data and/or reward functions.

Reinforcement learning based approaches, while extremely general, require good exploration of the state space to uncover high-reward behaviors. One of the most challenging modern problems in robotic learning is efficiently exploring high dimensional state spaces \cite{sutton, schmidhuber, silver}. Although many methods have been proposed for automatically exploring large state spaces — for example, epsilon-greedy methods \cite{sutton}, Boltzman exploration \cite{sutton2}, curiosity-driven methods \cite{schmidhuber2, pathak2}, Thompson sampling \cite{russo}, ensemble methods \cite{burda}, parameter exploration \cite{plappert, fortunato}, and self-imitation learning \cite{oh} — applying these methods in robotics remain very challenging due to sparse rewards and complex environment dynamics \cite{nair2}.

Learning from demonstrations requires collecting hundreds or thousands of examples demonstrating correct behavior \cite{zhang, rahmatizadeh, duan}. These expert demonstrations are used to train a model to imitate those behaviors. Datasets created by collecting expert demonstrations of individual tasks are usually limited in the number of visited states and the diversity of state-action trajectories. Typically these models are susceptible to compounding errors at test time if the agent encounters observations outside the narrow expert training distribution \cite{laskey}.

Recent works in the field of natural language processing suggest that using labeled task-specific datasets may no longer be necessary to achieve state-the-art results \cite{devlin, radford}. Using unlabeled text, BERT \cite{devlin} pre-trains a bidirectional network to predict a word conditioned on both left and right contexts. The model is then fine-tuned by supervised training of just one additional output layer. GPT-2 \cite{radford} uses a large non-specific dataset of web pages to train a general transformer-based algorithm \cite{vaswani} that predicts the next word conditioned on the previous context. The model achieves state-of-the-art performance on a variety of domain-specific tasks while not having been trained on any of the data specific to those tasks.

Continuing the trend of using self-supervised learning on non-specific datasets, learning control self-supervised from human teleoperated play \cite{lynch, gupta} was recently proposed as a means of acquiring a large repertoire of general goal-directed robotic behaviors. In this setting, a human operator teleoperates a robot, engaging in as many manipulations as they can think of in an environment, while recording the low level stream of robot observations and actions, directly addressing the hard exploration problem by covering the state space. This data is input to a simple hindsight relabeling technique, outputting a large number of goal-state conditioned imitation learning examples to power the learning of a single goal-directed policy. Lynch et al. \cite{lynch} show this process can be used to obtain a single agent that solves multiple user-specified visual manipulation tasks in zero shot, exceeding the performance of models trained on individual task expert demonstrations. In this work, we are interested in augmenting this process to provide large interaction-rich play datasets in the cheapest manner possible — by training a robot to collect them autonomously.

\section{PRELIMINARIES}

In this section we give background on goal-conditioned imitation learning, learning from play (LfP) and the motivation for the work presented here.

\subsection{Goal-conditioned Imitation Learning}

One of the simplest forms of learning from demonstrations is creating a model that can predict the best action $a_t$ given a current state $s_t$ \cite{argalla}. We train the model by optimizing an appropriate objective function, for example, maximizing the log likelihood. This approach to imitation learning leads to models that try to follow the state-action trajectories provided by the demonstrated examples. Intuitively, this is equivalent to taking the average of demonstrated actions at each specific state.

In goal-conditioned imitation learning \cite{kaelbling, lynch, ding} we would like to train policies that can reach an arbitrary state upon demand. Control is learned with a simple supervised learning objective: 

$$
\pazocal{L}_{GCIL}=\mathbb{E}_{(\tau, s_g)\sim D}\sum_{\tau}\log\pi(a_t|s_t,s_g) \eqno{(1)}
$$

\noindent
where $D$ is the set of demonstrations and $\tau$ is the set of state-actions pairs $(s_t, a_t)$ taken from an initial state to a goal state $s_g$, describing the task to complete.

\subsection{Learning from Play}
Intuitively, to have a goal conditioned policy that generalizes to a wide range of goals at test time, we need a training dataset that is broad and dense in its coverage of an environments available state space. That is, for every possible pair of current and goal states in the state space, the training should contain many different state-action sequences that connect them. Learning from play (Lynch et al. \cite{lynch}) takes this approach, scaling up goal-conditioned imitation learning to millions of diverse examples at training time mined from rich and unstructured teleoperated human play. The key insight here is that every short horizon window of states and actions mined from play contains an exact description of how the robot got from a particular initial state to a particular final state. This allows many windows to be sampled from play and relabeled as such in a self-supervised manner, creating a large dataset of ($\tau$, relabeled $s_g$) that can be used as input to goal conditioned imitation learning.

Human play can be both broad and dense. During play a human operator may imagine and teleoperate a robot to go from any initial state to any goal state. And the state-action trajectories taken by a human operator may be imaginative and diverse. Being broad and dense makes play datasets great for training general purpose control.

Concretely, LfP assumes access to a large dataset of hindsight-relabeled human teleoperated play $D_{play}$. This is used to train any goal-conditioned imitation learning architecture $\pi_{LfP}$ via simple goal-conditioned imitation learning. That is, in LfP (1) becomes

$$
\pazocal{L}_{LfP}=\mathbb{E}_{(\tau, s_g)\sim D_{play}}\sum_{\tau}\log\pi_{LfP}(a_t|s_t,s_g)\,. \eqno{(2)}
$$

In \cite{lynch} play data was generated by a human operator. The question that we would like to answer here is whether we can effectively multiply the size of human play by tens or hundreds by using autonomous play that is ``free" to generate. In the following sections we show that augmenting human play with robot play, as illustrated in Algorithm \ref{alg:play_cloning}, leads to substantially better self-supervised goal-conditioned control.

\begin{algorithm}[t]
\caption{Augmenting human play with robot play to learn self-supervised goal-conditioned control.}
  \begin{algorithmic}[1]
    \STATE {\bfseries Input:} $D_{play\_human} = \{(s_{0:t}, a_{0:t})^n\}^{\infty }_{n}$ \ \ \ \ \ // Dataset of high-dim observations and actions recorded during human teleoperation play.
    \STATE {\bfseries Input:} $D_{play\_robot} = \{\}$ \ \ \ \ \ // Empty dataset to hold cloned robot play.
    \STATE {\bfseries Input:} $\pi^{bc}_{\theta}(a_t | s_t)$ \ \ \ \ \ // Play cloning policy.
    \STATE {\bfseries Input:} $\pi^{lfp}_{\phi}(a_t | s_t, s_g)$ \ \ \ \ \ // Goal-conditioned imitation learning policy.
    \STATE {\bfseries Input:} Randomly initialize parameters $\{\theta,\ \phi\}$
    \FOR {$i=0 ... num\_train\_bc $}
    \STATE Sample batch from $D_{play\_human}$
    \STATE Compute $L\_clone$ \ \ \ \ \ // Learn to clone human play.
    \STATE Update parameters $\theta$ of $\pi^{bc}_{\theta}$
    \ENDFOR
    \FOR {$i=0 ... num\_play\_robot $}
    \STATE Sample play from $\pi^{bc}_{\theta}(a_t | s_t)$
    \STATE Add play to $D_{play\_robot}$ \ \ \ \ \ // Collect autonomous robot play.
    \ENDFOR
    \FOR {$i=0 ... num\_train\_lfp $}
    \STATE Sample batch from $D_{play\_human} + D_{play\_robot}$
    \STATE Compute $L\_lfp$ \ \ \ \ \ // Learn goal-conditioned control from human and robot play.
    \STATE Update parameters $\phi$ of $\pi^{lfp}_{\phi}$
    \ENDFOR
  \end{algorithmic}
\label{alg:play_cloning}
\end{algorithm}

\section{Learning To Play by Imitating Humans}

Here we describe the approach we use to imitate human play for the purposes of generating large amounts of free, manipulation-rich data that can be used to improve goal conditioned imitation learning.

The experiment environment, the process of collecting human play data, and control policy architectures were similar to those of \cite{lynch}, so we only provide a brief summary here. Fig. \ref{fig:play_cloning} provides a high-level overview of our approach. We start by collecting a relatively small quantity of human teleoperated play data, $D_{play}$. This data is used to train a stochastic \textit{play imitation policy} $\pi_{\theta}(a|s)$. We call this policy \textbf{Play-BC}. The goal is to learn a general mapping between states and a distribution over likely playful actions that a human would take in each state. In theory, we would be able to sample a large number of playful manipulation-rich behaviors from such a policy, generating new play interactions between the robot and play environment. Like original play, short-horizon windows taken from these autonomous play samples would also contain an exact description of how the agent got from some initial state to some final state, allowing us to append these to a large augmented training dataset for learning goal conditioned-control. The augmented data, that is, human play data supplemented by play imitation policy-generated data, can be used to train a downstream general purpose goal-conditioned policy as in \cite{lynch}. We carry this out in our experiments, evaluating the downstream policy on a set of 18 user-specified manipulation tasks \cite{lynch} to see if this improves goal-directed behavior. In the following sections we describe each step in detail.

\subsection{Collect Human Play}

Details of our play environment and the method for collecting human play are the same as \cite{lynch} — see https://learning-from-play.github.io for video examples of play. The play environment contains a simulated robot — consisting of arm and gripper — with eight degrees of freedom. The robot is situated in a realistic 3d living room, in front of a desk that has a rectangular block on it, three buttons that each control a separate light, a sliding door and a drawer. During play collection, a human operator teleoperates the robot to perform a variety of self-motivated tasks. For example, a single play episode may consist of an operator rolling the block to the right, opening and closing the drawer, rolling the block to the left, sliding the door to the left, and pressing a button.

Teleoperation is performed using a virtual reality headset that is worn by the human operator to perceive the environment. The human operator controls the robot using motion-tracked VR controllers that provide an intuitive way to manipulate and interact with the objects \cite{lynch, zhang}. We use an updated version of Mujoco HAPTIX system to collect the teleoperation data \cite{nair}.

\subsection{Training Play-BC}

The policy input is a vector of size nineteen. Eight of these coordinates define teleoperated robot states — six describe the state of the arm and two the gripper. The remaining eleven define the environment states. The position and orientation of the rectangular block is defined by six coordinates. The position of the drawer, sliding door and the buttons is describe by one coordinate each.

Actions are represented by eight coordinates. The change in the Cartesian position and orientation of the end effector of the robot arm is defined by six coordinates. Two coordinates define the change in gripper angles. Each coordinate is quantized into 256 bins. The policy predicts the discretized logistic distributions over these bins as described in \cite{salimans}.

In our experiments, we use the same neural network policy architecture as \cite{lynch}, implementing Play-BC as recurrent neural network that takes as input the low-dimension robot arm and environment states and predicts a mixture of logistic distributions (MODL) over discretized actions, where each action dimension has been discretized into 256 bins \cite{salimans}. Unless specified, in our experiments we use an RNN with 2 hidden layers and size of 2048 each. Our training examples consist of all short-horizon sampled windows of observations and actions from the unlabeled human play data \cite{lynch}, from window size 32 to 64. Since our data is collected at 30hz, this amounts to all roughly 1-2s windows from play.

\subsection{Generate Robot Play}

After we train a BC policy on unlabeled human play, we generate autonomous (cloned) play data by unrolling this policy in the play environment. We start by resetting the environment to a randomly selected initial state from play data and reset the RNN hidden state. We unroll the trained policy for a number of steps $n$ to create an autonomous play episode of duration $T$, for example, 1 minute. At each step, we predict a distribution over next action, sample an action, and apply it to the simulator, resulting in the next state. We repeat the process of resetting the environment, resetting the policy, and unrolling the policy to generate hours of cloned data.

\subsection{Train LfP Policy}

To validate whether play cloning helps skill learning, we combine the original human play and the cloned play and train a single LfP policy to convergence on the combined dataset. We then evaluate the trained LfP policy as in Lynch et. al \cite{lynch}, on a benchmark of 18 different object manipulation tasks, specified only at test time. While our method generates data for any goal-conditioned imitation learning architecture, we use the LMP architecture described in Lynch et. al \cite{lynch} which had the strongest performance previously in this setting.

\subsection{18-task Evaluation}

At test time, we evaluate the performance by measuring the average success rate of LfP policy on 18 user-specified manipulation tasks. These tasks are defined in the same play environment used for data collection and include grasp lift, grasp upright, grasp flat, drawer (to open a drawer), close drawer, open sliding, close sliding, knock object, sweep object, push red button, push green button, push blue button, put into shelf, pull out of shelf, rotate left, rotate right, sweep left, and sweep right. Detailed definitions of these tasks can be found in \cite{lynch}.

\section{EXPERIMENTS}

This section demonstrates that play cloning can be used as a way to scale up skill learning. We show that supplementing the human data with cloned data enhances state space coverage, an important property to many downstream robotic learning tasks. It also generalizes and substantially improves the performance of a goal-conditioned imitation learning policies, especially in the low-data regime.

\subsection{18-task performance}

\begin{figure}[t]
\centering
\includegraphics[width=0.48\textwidth]{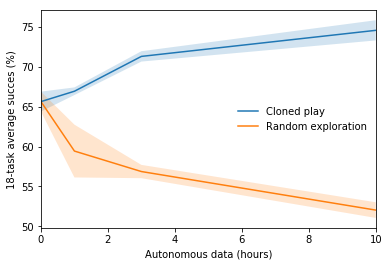}
\caption{18-task average success for an LfP agent trained on 32 minutes of human play data and different amounts of autonomously generated data. Supplementing human play data with 10 hours of BC cloned data substantially improved agent average success. The graph also shows 18-task average success for the same policy trained on 32 minutes of human play data and different amounts of random exploration episodes, used here as a baseline.}
\label{fig:success_vs_hours}
\end{figure}

In our experiments we collected and used 32 minutes of human play data. A Play-BC policy was trained on this data and unrolled to generate hours of cloned play data. We then trained an LfP policy on a training dataset comprised of the original 32 minutes of human play data and different quantities of cloned play data. We evaluated the performance of the LfP policy by measuring average success on the 18 tasks. Fig. \ref{fig:success_vs_hours} shows the 18-task average success comparison. A single LfP policy trained only on 32 minutes of human play data achieved an 18-task average success of $65.6\% \pm 1.2\%$ whereas one trained on 32 minutes of human play data and 10 hours of cloned data achieved an average success of $74.6\% \pm 1.3\%$. This is a substantial $13.7\%$ improvement in performance. The cloned data was generated with a Play-BC policy having an RNN with 2 layers and 2048 elements each and the length of each cloned play episode was 1 minute.

Fig. \ref{fig:success_vs_hours} also shows the effect of augmenting human play data with random exploration data. We can see that the performance of the model significantly degrades with adding more random exploration episodes to the training data. Data created by random exploration are insufficiently rich. Many complex object manipulations, such as grasping objects off of a table, opening and closing doors, etc. require precise control and are unlikely to happen in the course of random action sampling. To generate random exploration episodes, at each step we sampled the actions from a normal distribution with the same mean and standard deviation as the human play data. We clipped the actions to a maximum and minimum value to prevent the robot from making flailing motions. The random exploration episodes each had a duration of 1 minute.

Because we are using supervised learning to generate cloned plays, we use a high capacity BC policy with 2 layers and 2048 elements each that can capture subtle details of human play behavior. Table \ref{table:suboptimal_cloning} shows the effect of using cloned play data that were generated with BC policies with lower capacities. As expected, using clones generated with Play-BC policies with lower capacities lead to lower 18-task average success. For example, when we trained the same LfP policy on 32 minutes of human data and 10 hours of clone data generated by a BC policy of 1 layer and 1024 elements, the average success was $54.2\%$. A high capacity Play-BC policy generates a more diverse range of behaviors, leading to richer cloned plays.

\begin{table}[t]
\caption{Effect of cloning with sub-optimal Play-BC policy}
\label{table:suboptimal_cloning}
\begin{center}
\begin{tabular}{|c|c||c|}
\hline
Number of layers & Size of layer & 18-task average success (\%)\\
\hline\hline
2 & 2048 & $74.6 \pm 1.3$\\
\hline
2 & 1024 & $60.3 \pm 1.3$\\
\hline
1 & 2048 & $56.7 \pm 1.2$\\
\hline
1 & 1024 & $54.2 \pm 1.4$\\
\hline
\end{tabular}
\end{center}
\end{table}

\begin{figure}[t]
\centering
\includegraphics[width=0.48\textwidth]{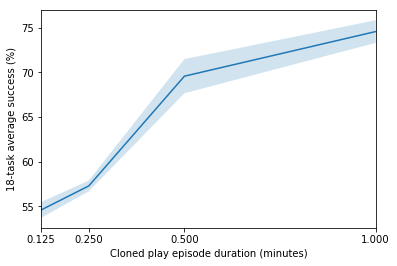}
\caption{18-task average success of LfP policy as a function of duration of BC imitated play episodes. LfP policy was trained on 32 minutes of human play data and 10 hours of imitated play data. Clone length (duration) is how long an agent continued behavioral cloning before it was reset.}
\label{fig:success_vs_clone_length}
\end{figure}

We also explored the effect of the length (duration) of cloned play data. That is, whether resetting the Play-BC policy too often (that is, shorter imitated play episodes) have a result on the 18-task average success. Fig. \ref{fig:success_vs_clone_length} shows that longer autonomous play episodes result in better performance. Our suspicion is that in longer episodes the cloning agent has a wider range of behaviors, visiting states farther from the ones seen in the human training data, leading to a more uniform exploration and state space coverage.

\subsection{State Space Coverage}

Play-BC generated data improves the average success of the LfP agent by introducing data that covers new areas of the state space. To obtain a qualitative measure of state space coverage, we quantize the state space to discrete bins and count the number of bins that are covered by the data.

\begin{figure}[t]
\centering
\includegraphics[width=0.48\textwidth]{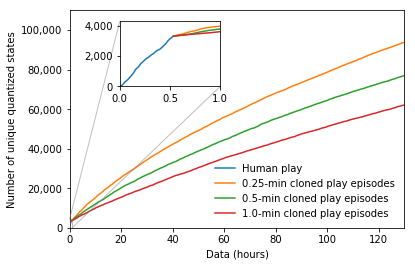}
\caption{Number of unique quantized states in the data used to train LfP policy. The first 32 minutes correspond to the number of unique quantized states covered by the human training data. The remaining correspond to the data generated by play behavioral cloning. The inset shows a zoomed in version of the first hour of data. The rate of unique quantized states covered per hour of data is higher when the agent is reset more often during play cloning. The rate of unique quantized states covered per hour by human data is 10-16 times higher than that of Play-BC cloned data.}
\label{fig:num_unique_quantized_states}
\end{figure}

For this analysis we exclude robot states and consider only the environment states, that is, the state of the rectangular block, buttons, drawer and sliding door. We quantize each environment state into ten bins by equally dividing the difference between the minimum and maximum value of a state in the human training data into eight bins and leaving two bins for values less than minimum or greater than maximum. Therefore our state space is quantized into $10^{11}$ bins. We count the number of unique bins visited by the agent as a measure of state space coverage. Fig. \ref{fig:num_unique_quantized_states} shows cumulative number of unique quantized states for 32 minutes of human training superimposed by tens of hours of cloned play data. The first 32 minutes correspond to the human play data and the remaining correspond to the cloned play data drawn on top of the human data to represent the new unique states visited by the cloned play data.

Fig. \ref{fig:num_unique_quantized_states} shows that autonomous play data generated by Play-BC policy introduces new states beyond what is in the human play data and the number of unique states visited by the BC policy increases almost linearly for tens of hours of data. The inset shows the first hour of data zoomed in. The first 32 minutes (0.53 hours) correspond to the human data. The remaining correspond to Play-BC generated data with different episode duration. Inset shows that the slope of the human play data is steeper than that of cloned play data, indicating that the rate of visiting new unique states is slower for the Play-BC policy compared to a human operator.

Fig \ref{fig:num_unique_quantized_states} also compares the state space coverage for clones of different episode lengths. The same quantity of cloned data with shorter episodes has a higher state space coverage than that of longer episodes. As we saw previously in Fig. \ref{fig:success_vs_clone_length} an LfP policy trained on cloned play data with longer episodes has a significantly better 18-task average success than that of shorter episodes. Intuitively this makes sense because longer clones have a higher chance of exhibiting behavior that is more different than the behaviors available in the human training data, leading to a policy that generalizes better. This also suggests that better heuristics should be developed. That is, simple state space coverage count does not provide a reliable measure for predicting the richness of the data and the performance of an LfP policy trained on the data. .

\subsection{Qualitative Examples}
Fig. \ref{fig:cloned_play_sequence} shows qualitative examples of autonomous play. Video examples can be found \href{\ltpvideos/undirected_play1.mp4}{here}, \href{\ltpvideos/undirected_play2_20s.mp4}{here}, \href{\ltpvideos/undirected_play3_20s.mp4}{here} and \href{\ltpvideos/undirected_play4_20s.mp4}{here}.
We also present failure cases \href{\ltpvideos/failures/undirected_play_failure1_20s.mp4}{here} and \href{\ltpvideos/failures/undirected_play_failure2_20s.mp4}{here} where the play drifts out of the frame or remains rather static and does not exhibit useful or functional behavior.

\section{CONCLUSIONS}

Our work demonstrates that by imitating human play we can autonomously generate rich task-agnostic interactions that can be used as additional training data to substantially improve the performance of an LfP policy. This can pave the way for creating robots that can learn to adapt to new environments via imitating a modest quantity of human play in such environments. Some direct extensions of this work include using more complex BC policies — for example, using stochastic models with latent plans — to generate a wider variety of imitated play, and creating cloned play data with longer episodes. NLP methods such as BERT and GPT-2 use non-specific datasets to train networks tasked to predict a word based on the context. Similarly, training on non-specific play data LfP uses state history conditioned on goal state to predict actions. Inspired by the importance of using large datasets in NLP to achieve state-of-the-art performance, one future research would be to autonomously generate much larger play imitation datasets to explore whether we can train LfP models with higher capacity to achieve human-like performance.

\addtolength{\textheight}{-6cm}  % This command serves to balance the column lengths on the last page of the document manually. It shortens the textheight of the last page by a suitable amount. This command does not take effect until the next page so it should come on the page before the last. Make sure that you do not shorten the textheight too much.

%%%%%%%%%%%%%%%%%%%%%%%%%%%%%%%%%%%%%%%%%%%%%%%%%%%%%%%%%%%%%%%%%%%%%%%%%%%%%%%%

%%%%%%%%%%%%%%%%%%%%%%%%%%%%%%%%%%%%%%%%%%%%%%%%%%%%%%%%%%%%%%%%%%%%%%%%%%%%%%%%

\end{document}